\documentclass{article} % For LaTeX2e
\usepackage{nips10submit_e,times}
\usepackage{graphicx} % more modern
\usepackage{subfigure} 
\usepackage{amsmath, amsthm}
\usepackage{amsfonts}
\usepackage{mlapa}

\title{Adaptive Parallel Tempering for Stochastic Maximum Likelihood Learning of RBMs}

\author{
Guillaume Desjardins, Aaron Courville and Yoshua Bengio\\
Dept. IRO\\
Universit\'e de Montr\'eal\\
Montr\'eal, QC  \\
\texttt{\{desjagui,courvila,bengioy\}@iro.umontreal.ca}
}

\nipsfinalcopy % Uncomment for camera-ready version

\newcommand{\E}{\mathbb{E}}
\newcommand{\Egy}{{\bf E}}
\newcommand{\x}{\mathbf{x}}
\newcommand{\h}{\mathbf{h}}
\newcommand{\vis}{\mathbf{v}}
\newcommand{\nv}{\mathbf{v^{{\bf -}}}}
\newcommand{\nh}{\mathbf{h^{{\bf -}}}}

\renewcommand{\b}{\mathbf{b}}
\renewcommand{\c}{\mathbf{c}}
\newcommand{\W}{\mathbf{W}}
\newcommand{\T}{{\bf \mathcal T}}

\begin{document}

\maketitle

\begin{abstract}
Restricted Boltzmann Machines (RBM) have attracted a lot of attention of late, as one the principle building blocks of deep networks. Training RBMs remains problematic however, because of the intractibility of their partition function. The maximum likelihood gradient requires a very robust sampler which can accurately sample from the model despite the loss of ergodicity often incurred during learning. While using Parallel Tempering in the negative phase of Stochastic Maximum Likelihood (SML-PT) helps address the issue, it imposes a trade-off between computational complexity and high ergodicity, and requires careful hand-tuning of the temperatures. In this paper, we show that this trade-off is unnecessary. The choice of optimal temperatures can be automated by minimizing average return time (a concept first proposed by \cite{Katzgraber06}) while chains can be spawned dynamically, as needed, thus minimizing the computational overhead. We show on a synthetic dataset, that this results in better likelihood scores.
\end{abstract}

\section{Introduction}

Restricted Boltzmann Machines (RBM)~\cite{Freund+Haussler-94,Welling05,Hinton06} have become a model of choice for learning
unsupervised features for use in deep feed-forward architectures \cite{Hinton06,Bengio-2009} as well
as for modeling complex, high-dimensional distributions \cite{Welling05,TaylorHintonICML2009,Larochelle+al-2010}. 
Their success can be explained in part through the bi-partite structure of their
graphical model. Units are grouped into a visible layer $\vis$ and a hidden layer
$\h$, prohibiting connections within the same layer. The use of latent
variables affords RBMs a rich modeling capacity, while the conditional
independence property yields a trivial inference procedure. 

RBMs are parametrized by an energy function $\Egy(\vis,\h)$ which is converted to
probability through the Boltzmann distribution, after marginalizing out the
hidden units. The probability of a given configuration $p(\vis)$ is thus given by
$p(\vis) = \frac{1}{Z} \sum_{\h} \exp(-\Egy(\vis,\h))$, where $Z$ is the partition function
defined as $Z=\sum_{\vis,\h} \exp(-\Egy(\vis,\h))$.

Despite their popularity, direct learning of these models through maximum
likelihood remains problematic. The maximum likelihood gradient
with respect to the parameters $\theta$ of the model is:
\begin{eqnarray}
\label{eq:maxll}
\frac{\partial \log p(\vis)}{\partial \theta} 
&=& -\sum_{\h} p(\h|\vis) \frac{\partial \Egy(\vis,\h)}{\partial \theta}
+ \sum_{\nv,\nh} p(\nv,\nh) \frac{\partial \Egy(\nv,\nh)}{\partial \theta}
\label{eq:pos-neg}
\end{eqnarray}

The first term is trivial to calculate and is referred to as the {\bf positive
phase}, as it raises the probability of training data. The second term or {\bf
negative phase} is intractable in most applications of interest, as it involves an
expectation over $p(\vis,\h)$. Many learning algorithms have been
proposed in the literature to address this issue:

\begin{itemize}
%===== CD =====%
\item Contrastive Divergence (CD)~\cite{Hinton99,Hinton2002} replaces the expectation with
a finite set of negative samples, which are obtained by running a short Markov
chain initialized at positive training examples. This yields a biased, but
low-variance gradient which has been shown to work well as a feature extractor
for deep networks such as the Deep Belief Network~\cite{Hinton06}.
%===== PCD =====%
\item Stochastic Maximum Likelihood (SML) or Persistent Contrastive Divergence (PCD)
\cite{Younes98onthe,Tieleman08} on the other hand, relies on a persistent Markov
chain to sample the negative particles. The chain is run for a small number of
steps between consecutive model updates, with the assumption that the Markov
chain will stay close to its equilibrium distribution as the parameters evolve.
Learning actually encourages this process, in what is called the
``fast-weight effect'' \cite{TielemanT2009}.
%====== RM / SM ======%
\item Ratio Matching and Score Matching~\cite{Hyvarinen-2005,Hyvarinen-2007} 
avoid the issue of the partition function
altogether by replacing maximum likelihood by another learning principle, based
on matching the change in likelihood to that implied by the empirical distribution.
\end{itemize}

\cite{Marlin10InductivePrinciples} recently compared these algorithms on a
variety of tasks and found SML to be the most attractive method when taking
computational complexity into account.
Unfortunately, these results fail to address the main shortcomings of SML.
First, it relies on Gibbs sampling to extract negative samples: a poor choice
when sampling from multi-modal distributions. Second, to guarantee convergence,
the learning rate must be annealed throughout learning in order to offset the
% TODO: define ergodicity
loss of ergodicity 
\footnote{We use the term ``ergodicity'' rather loosely, to reflect the amount
of time required for the states sampled by the Markov chain, to reflect the
true expectation we wish to measure.}
incurred by the Markov chain due to parameter updates
\cite{Younes98onthe,Desjardins+al-2010}.
Using tempering in the negative phase of
SML~\cite{Desjardins+al-2010,Cho10IJCNN,Salakhutdinov-2010,Salakhutdinov-ICML2010}
appears to address these issues to some extent. By performing a random
walk in the joint (configuration, temperature) space, negative particles can
escape regions of high probability and travel between disconnected modes. Also,
since high temperature chains are inherently more ergodic, the sampler as a
whole exhibits better mixing and results in better convergence properties than
traditional SML.

Tempering is still no panacea however. Great care must be taken to select the
set of temperatures $\T=\{T_1,...,T_M ; T_1 < T_i < T_M \ \forall i \in [1,M], M
\in \mathbb{N}\}$ over which to run the simulation.  Having too few or incorrectly
spaced chains can result in high rejection ratios (tempered transition), low
return rates (simulated tempering) or low swap rates between neighboring chains
(parallel tempering), which all undermine the usefulness of the method. In this
work, we show that the choice of $\T$ can be automated for parallel tempering,
both in terms of optimal temperature spacing, as well as the number of chains
to simulate. Our algorithm relies heavily on the work of \cite{Katzgraber06},
who were the first to show that optimal temperature spacing can be obtained
by minimizing the average {\bf return time} of particles under simulation. 
%This is defined as the number of steps required for a particle to perform a round-trip between $T_1$ and $T_M$.

The paper is organized as follows. We start with a brief review of SML, then
explore the details behind SML with Parallel Tempering (SML-PT) as described in
\cite{Desjardins+al-2010}. Following this, we show how the algorithm of
Katzgraber et al. can be adapted to the online gradient setting for use with
SML-PT and show how chains can be created dynamically, so as to maintain a
given level of ergodicity throughout training. We then proceed to show various
results on a complex synthetic dataset.

% Things to touch n
% * motivation: why is learning RBMs so important
% * cite inductive principles paper
% * cover previous work on tempering PCD
% * explain what is novel about this paper
% * give outline of paper

\section{SML with Optimized Parallel Tempering}

\subsection{Parallel Tempered SML (SML-PT)}

We start with a very brief review of SML, which will serve mostly to anchor our
notation. For details on the actual algorithm, we refer the interested reader to 
\cite{TielemanT2009,Marlin10InductivePrinciples}.
RBMs are parametrized by $\theta=\{\W, \b, \c\}$, 
where $\b_i$ is the $i$-th hidden bias,
$\c_j$ the $j$-th visible bias and
$W_{ij}$ is the weight connecting units $h_i$ to $v_j$. 
They belong to the family of log-linear models whose energy function is given
by $\Egy(\x) = -\sum_k \theta_k \phi_k(\x)$, where $\phi_k$ are 
functions associated with each parameter $\theta_k$. In the case of RBMs, $\x=(\vis,\h)$ and
$\phi(\vis,\h) = (\h\vis^T, \h, \vis)$. For this family of model, the gradient of
Equation~\ref{eq:maxll} simplifies to:
\begin{eqnarray}
\label{eq:maxll2}
\frac{\partial \log p(\vis)}{\partial \theta} 
&=& \E_{p(\h|\vis)}[\phi(\vis,\h)] - \E_{p(\vis,\h)}[\phi(\vis,\h)].
\end{eqnarray}

As was mentioned previously, SML approximates the gradient by drawing negative
phase samples (i.e. to estimate the second expectation) 
from a persistent Markov chain, which attempts to track changes
in the model. If we denote the state of this chain at timestep $t$ as
$\vis^{-}_t$ and the i-th training example as $\vis^{(i)}$, then the stochastic
gradient update follows $\phi(\vis^{(i)},\tilde{\h}) - \phi(\tilde{\vis}^{-}_{t+k},
\tilde{\h}^{-}_{t+k})$, where $\tilde{\h} = \E[\h|\vis=\vis^{(i)}]$, and
$\tilde{\vis}^{-}_{t+k}$ is obtained after $k$ steps of alternating Gibbs
starting from state $\vis^{-}_t$ and $\tilde{\h}^{-}_{t+k} =
\E[\h|\vis=\vis^{-}_{t+k}]$.

\smallskip
Training an RBM using SML-PT maintains the positive phase as is.  During the
negative phase however, we create and sample from an an extended set of $M$
persistent chains, $\{p_{\beta_i}(\vis,\h)| i \in [1,M], \beta_i \ge \beta_j \iff i < j\}$. 
Here each $p_{\beta_i}(\vis,\h) = \frac{\exp(-\beta_i \Egy(\x))}{Z(\beta_i)}$
represents a smoothed version of the distribution we wish to sample from, with
the inverse temperature $\beta_i = 1/T_i \in [0,1]$ controlling the degree of smoothing.
Distributions with small $\beta$ values are easier to sample from as they
exhibit greater ergodicity.

After performing $k$ Gibbs steps for each of the $M$ intermediate distributions,
cross-temperature state swaps are proposed between neighboring chains using a
Metropolis-Hastings-based swap acceptance criterion. If we denote by $\x_i$ the joint state (visible
and hidden) of the $i$-th chain, the swap acceptance ratio $r_i$ for swapping
chains ($i$,$i+1$) is given by:
\begin{align}
    r_i &= \max(1, \frac {p_{\beta_i}(\x_{i+1})p_{\beta_{i+1}}(\x_i)} 
                     {p_{\beta_i}(\x_i)p_{\beta_{i+1}}(\x_{i+1})})
\end{align}

Although one might reduce variance by using free-energies to compute swap
ratios, we prefer using energies as the above factorizes nicely into the
following expression:
\begin{align}
r_i = \exp((\beta_i - \beta_{i+1}) \cdot (\Egy(\x_i) - \Egy(\x_{i+1}))),
\end{align}

While many swapping schedules are possible, we use the Deterministic Even Odd
algorithm (DEO) \cite{Lingenheil200980}, described below.

\subsection{Return Time and Optimal Temperatures}
\label{sec:returntime}

Conventional wisdom for choosing the optimal set $\T$ has relied on the
``flat histogram'' method which selects the parameters $\beta_i$ such that
the pair-wise swap ratio $r_i$ is constant and independent of the index $i$.
Under certain conditions (such as when sampling from multi-variate Gaussian
distributions), this can lead to a geometric spacing of the temperature
parameters \cite{Neal94b}. \cite{Behrens2010} has recently shown that geometric
spacing is actually optimal for a wider family of distributions characterized by
$\E_{\beta}(\Egy(x)) = K_1/\beta + K_2$, where $\E_\beta$ denotes
the expectation over inverse temperature and $K_1,K_2$ are arbitrary
constants. 

%% TODO: TRIPLE-CHECK THIS
Since this is clearly not the case for RBMs, we turn to the work of
\cite{Katzgraber06} who propose a novel measure for optimizing $\T$.
Their algorithm directly maximizes the ergodicity of the sampler by minimizing
the time taken for a particle to perform a round-trip between $\beta_1$ and
$\beta_M$. This is defined as the average ``return time'' $\tau_{rt}$. The
benefit of their method is striking: temperatures automatically pool around
phase transitions, causing spikes in local exchange rates and maximizing the
``flow'' of particles in temperature space.

The algorithm works as follows. For $N_s$ sampling updates:
\begin{itemize}
\item assign a label to each particle: those swapped into $\beta_1$ are labeled
      as ``up'' particles. Similarly, any ``up'' particle swapped into $\beta_M$
      becomes a ``down'' particle.
\item after each swap proposal, update the histograms $n_{u}(i), n_{d}(i)$,
      counting the number of ``up'' and ``down'' particles for the Markov chain associated with $\beta_i$.
  \item define $f_{up}(i) = \frac {n_{u}(i)} {n_{u}(i) + n_{d}(i)}$, the fraction of
      ``up''-moving particles at $\beta_i$. By construction, notice that
      $f_{up}(\beta_1)=1$ and $f_{up}(\beta_M) = 0$. $f_{up}$ thus defines a
      probability distribution of ``up'' particles in the range $[\beta_1,\beta_M]$.
\item The new inverse temperature parameters $\beta'$ are chosen as the ordered
    set which assigns equal probability mass to each chain. This yields
    an $f_{up}$ curve which is linear in the chain index.
\end{itemize}

The above procedure is applied iteratively, each time increasing $N_s$ so as to
fine-tune the $\beta_i$'s. To monitor return time, we can simply maintain a
counter $\tau_i$ for each particle $\x_i$, which is (1) incremented at every
sampling iteration and (2) reset to 0 whenever $\x_i$ has label ``down'' and is
swapped into $\beta_1$. A lower-bound for return time is then given by 
${\hat{\tau}_{rt}} = \sum_{i=0}^M \tau_i$.

% if sufficient amount of time/space left, discuss the equivalence of flat
% histogram method and minizing return time in the case of distributions where,
% at a given energy level E, the configuration space can be sampled easily. We
% are unfortunately dealing with the "broken ergodicity". In this setting
% measuring average swap rates doesn't even make sense.
% \cite{Nadler07} 

\subsection{Optimizing $\T$ while Learning}

\subsubsection{Online Beta Adaptation}

While the above algorithm exhibits the right properties, it is not very well
suited to the context of learning. When training an RBM, the distribution we are
sampling from is continuously changing. As such, one would expect the optimal
set $\T$ to evolve over time. We also do not have the luxury
of performing $N_s$ sampling steps after each gradient update. 

Our solution is simple: the histograms $n_u$ and $n_d$ are updated
using an exponential moving average, whose time constant is in the order of the
return time $\hat{\tau}_{rt}$. Using $\hat{\tau}_{rt}$ as the time constant is crucial as
it allows us to maintain flow statistics at the proper timescale. If an ``up''
particle reaches the $i$-th chain, we update $n_u(i)$ as follows:
\begin{align}
    n_u^{t+1}(i) = n_u^{t}(i) (1-1/{\hat{\tau}_{rt}^t}) +  1/{\hat{\tau}_{rt}^t},
\end{align}
where $\hat{\tau}_{rt}^t$ is the estimated return time at time $t$.

Using the above, we can estimate the set
of optimal inverse temperatures $\beta_i'$. Beta values are updated by
performing a step in the direction of the optimal value: 
$\beta_i^{t+1} = \beta_i^t + \mu (\beta_i' - \beta_i^t)$, where $\mu$ is a
learning rate on $\beta$. The properties of \cite{Katzgraber06} naturally
enforce the ordering constraint on the $\beta_i$'s.

\subsubsection{Choosing $M$ and $\beta_M$}

Another important point is that \cite{Katzgraber06} optimizes the set
$\T$ while keeping the bounds $\beta_1$ and $\beta_M$ fixed. While 
$\beta_1=1$ is a natural choice, we expect the optimal $\beta_M$ to vary during
learning. For this reason, we err on the side of caution and use $\beta_M=0$,
relying on a {\bf chain spawning process} to maintain sufficiently high swap
rates between neighboring parallel chains. Spawning chains as required by the
sampler should therefore result in increased stability, as well as computational
savings.

\cite{Lingenheil200980} performed an interesting study where they compared the
round trip rate $1/\tau_{rt}$ to the average swap rate measured across all
chains. They found that the DEO algorithm, which alternates between proposing
swaps between chains $\{(i,i+1); \forall\, {\rm even}\, i\}$ followed by $\{(i,i+1);
\forall\, {\rm odd}\, i\}$), gave rise to a concave function with a broad maximum around
an average swap rate of $\bar{r} = \sum_i{r_i} \approx 0.4$

Our temperature adaptation therefore works in two phases:
\begin{enumerate}
    \item The algorithm of Katzgraber et. al is used to optimize 
        $\{\beta_i; 1 < i < M\}$, for a fixed M.
    \item Periodically, a chain is spawned whenever $\bar{r} < \bar{r}_{min}$, 
          a hyper-parameter of the algorithm.
\end{enumerate}

Empirically, we have observed increased stability when the index $j$ of the new
chain is selected such that $j=\mathrm{argmax}_i (|f_{up}(i) - f_{up}(i+1)|)$,
$i \in [1,M-1]$. To avoid a long burn-in period, we initialize the new chain
with the state of the $(j+1)$-th chain and choose its inverse temperature as
the mean $(\beta_j + \beta_{j+1})/2$. A small but fixed burn-in period allows
the system to adapt to the new configuration.

\section{Results and Discussion}

We evaluate our adaptive SML-PT algorithm (SML-APT) on a
complex, synthetic dataset. This dataset is heavily inspired from the one used
in \cite{Desjardins+al-2010} and was specifically crafted to push the limits
of the algorithm. 

It is an online dataset of $28$x$28$ binary images, where each example is
sampled from a mixture model with probability density function $f_{X}(x) =
\sum_{m=1}^{5} w_m f_{Y_m}(x)$. Our dataset thus consists of $5$ mixture
components whose weights $w_m$ are sampled uniformly in the unit interval and
normalized to one. Each mixture component $Y_m$ is itself a random $28$x$28$
binary image, whose pixels are independent random variables having a probability
$p_m$ of being flipped. 
%small $p_m$ is akin to the variance parameter of Gaussian distributions. 
From the point of view of a sampler performing a
random walk in image space, $p_m$ is inversely proportional to the difficulty
of finding the mode in question. The complexity of our synthetic dataset comes
from our particular choice of $w_m$ and $p_m$.
\footnote{$w=[0.3314, 0.2262, 0.0812, 0.0254, 0.3358]$ and $p=[0.0001, 0.0137, 0.0215, 0.0223, 0.0544]$}
Large $w_m$ and small $p_m$ lead
to modes which are difficult to sample and in which a Gibbs sampler would tend
to get trapped. Large $p_m$ values on the other hand will tend to intercept "down" moving particles and thus present a challenge for parallel tempering.

\begin{figure}[ht]
    \centering
    \hspace*{-1.0cm}
    \subfigure[Log-likelihood]
    {
        \label{fig:fig11}
        \includegraphics[scale=0.4]{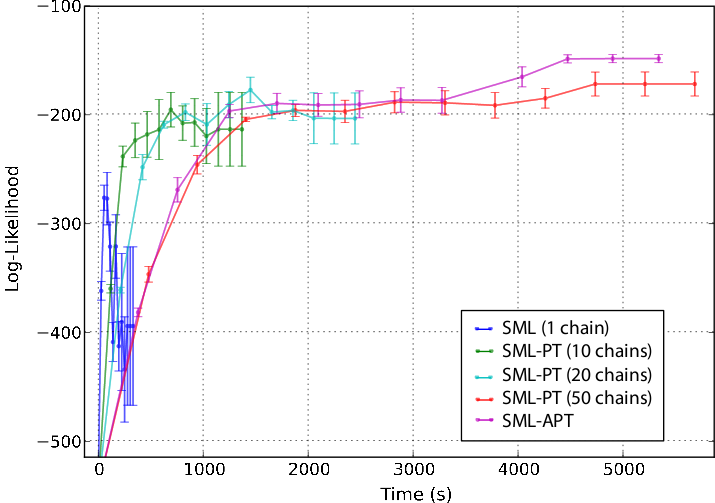}
    }
    \hspace*{-.2cm}
    \subfigure[Return Time]
    {
        \label{fig:rtime}
        \includegraphics[scale=0.38]{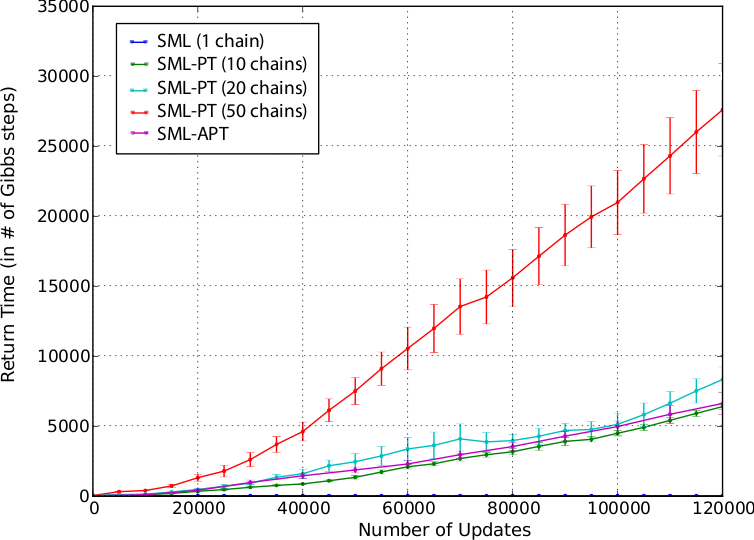}
    }
    \hspace*{-.5cm}
    \caption[] {(a) Comparison of training likelihood as a function of time for
    standard SML, SML-PT with 10/20/50 chains and the proposed SML-APT
    (initialized with 10 chains). SML-APT adapts the temperature set $\T =
    \{T_1,...,T_M ; T_1 < T_i < T_M \}$ to minimize round
    trip time between chains $T_1$ and $T_M$, and modifies the number of chains
    $M$ to maintain a minimal average swap rate. The resulting sampler exhibits
    greater ergodicity and yields better likelihood scores than standard SML
    and SML-PT, without requiring a careful hand-tuning of $\T$.
    (b) Average return time of each algorithm. SML-APT successfully minimizes
    this metric resulting in a return time similar to SML-PT 10, while still
    outperforming SML-PT $50$ in terms of likelihood.
    Errors bars represent standard error on the mean.
    \label{fig:fig1}}
\end{figure}

Figure~\ref{fig:fig11} compares the results of training a $10$ hidden unit RBM, using
standard SML, SML-PT with $\{10,20,50\}$ parallel chains and our new SML-APT
algorithm. We performed $10^5$ updates (followed by $2 \cdot 10^4$ steps of
sampling) with mini-batches of size 5 and tested learning rates in
$\{10^{-3},10^{-4}\}$, $\beta$ learning rates in $\{10^{-3},10^{-4},10^{-5}\}$. For each
algorithm, we show the results for the best performing hyper-parameters,
averaging over $5$ different runs. Results are plotted with respect to
computation time to show the relative computational cost of each algorithm.

% results take from:
% /data/lisa6/desjagui/smlpt/nips10_workshop/exactll/id9/subset
% plotresults -dir . -type hdf5 -xnode nll -x t -y nll -fname outputs.h5 -filter rbm.n_beta spawn_beta adapt_beta
 
\begin{figure}[ht]
    \centering
    \hspace*{-0.7cm}
    \subfigure[SML-APT]
    {
        \label{fig:fup1}
        \includegraphics[scale=0.35]{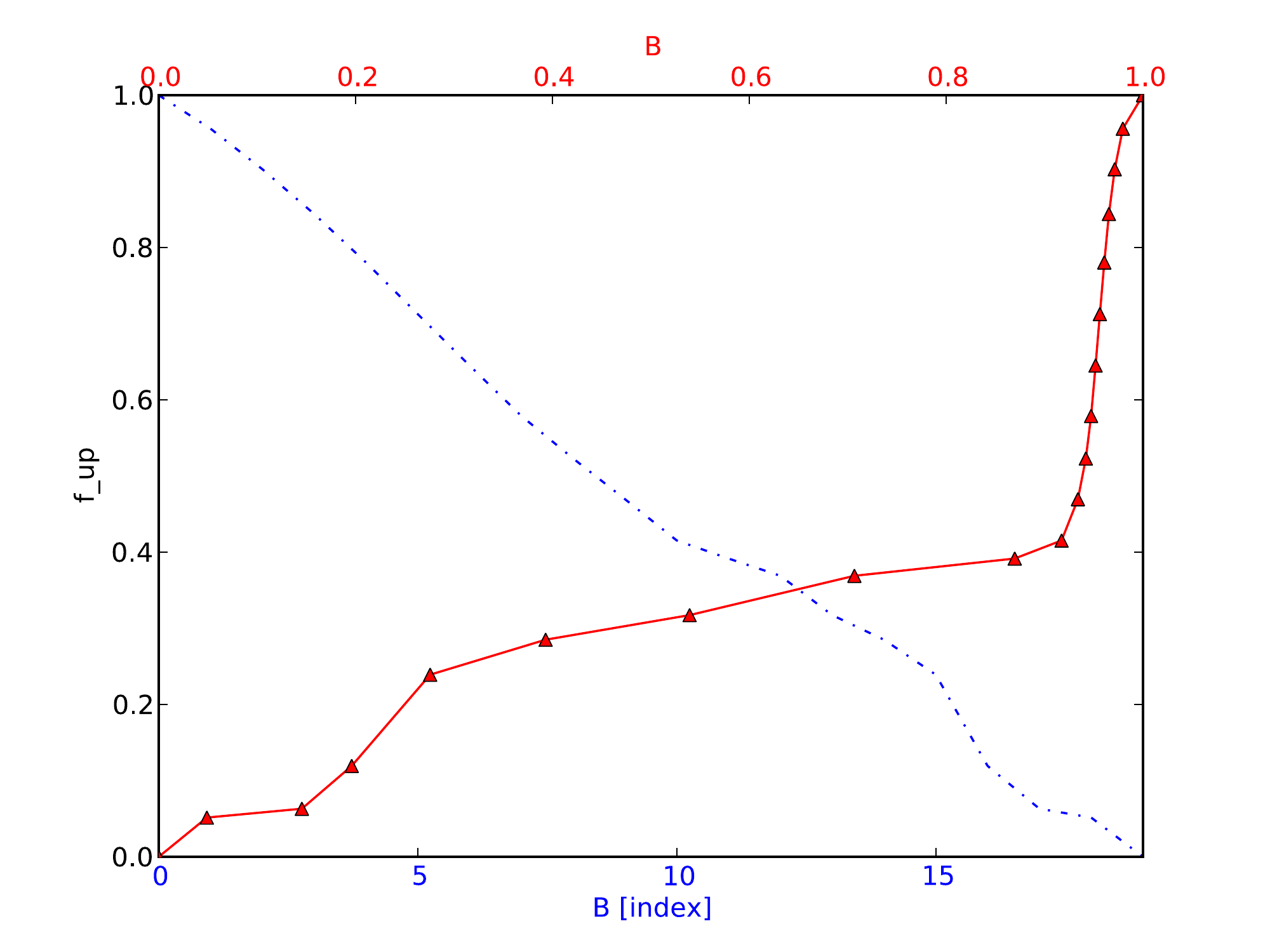}
    }
    \hspace*{-.5cm}
    \subfigure[SML-PT 50]
    {
        \label{fig:fup2}
        \includegraphics[scale=0.35]{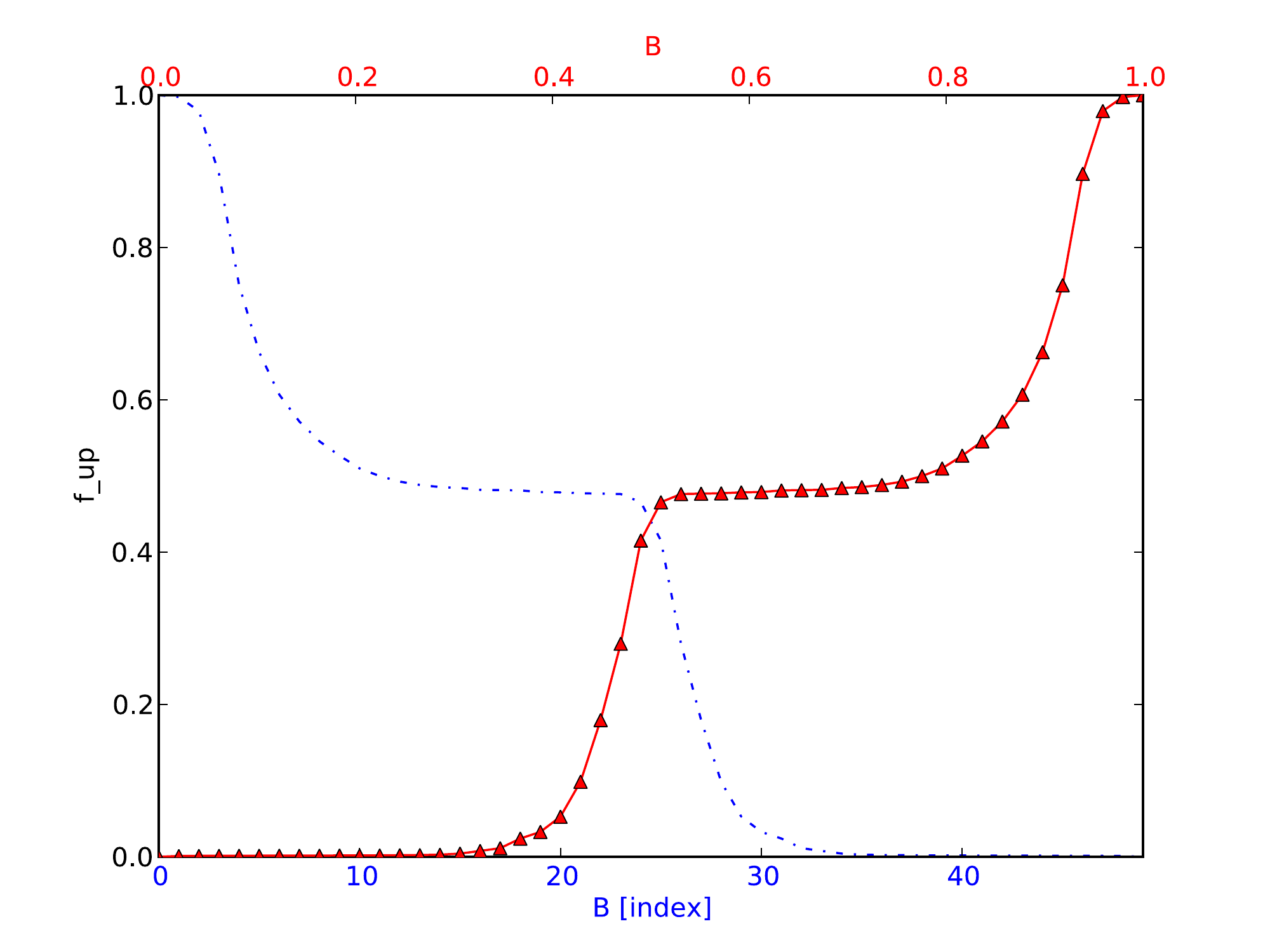}
    }
    \hspace*{-.5cm}
    \caption[] {
    Return time is minimized by tagging each particle with a label: ``up'' if
    the particle visited $T_1$ more recently than $T_M$ and ``down'' otherwise.
    Histograms $n_u(i)$ and $n_d(i)$ track the number of up/down particles at
    each temperature $T_i$. Temperatures are modified such that the ratio
    $f_{up}(i) = n_u(i) / (n_u(i) + n_d(i))$ is linear in the index $i$.
    (a) $f_{up}$ curve obtained with SML-APT, as a function of temperature
    index (blue) and inverse temperature (red). SML-APT achieves a linear
    $f_{up}$ in the temperature index $i$.
    (b) Typical $f_{up}$ curve obtained with SML-PT (here using 50
    chains). $f_{up}$ is not linear in the index $i$, which translates to
    larger return times as shown in Fig.~\ref{fig:rtime}.
    \label{fig:fup}}
\end{figure}

As we can see, standard SML fails to learn anything meaningful: the Gibbs
sampler is unable to cope with the loss in ergodicity and the model diverges.
SML-PT on the other hand performs much better. Using more parallel chains in
SML-PT consistently yields a better likelihood score, as well as reduced
variance. This seems to confirm that using more parallel chains in SML-PT increases the
ergodicity of the sampler. Finally, SML-APT outperforms all other methods. As
we will see in Figure~\ref{fig:fup}, it does so using only $20$ parallel
chains. Unfortunately, the computational cost seems similar to 50 parallel
chains. We hope this can be reduced to the same cost as SML-PT with $20$ chains
in the near future.
Also interesting to note, while the variance of all methods increase with
training time, SML-APT seems immune to this issue.

We now compare the various metrics being optimized by our adaptive algorithm.
Figure~\ref{fig:rtime} shows the average return time for each of the
algorithms. We can see that SML-APT achieves a return time which is comparable
to SML-PT with 10 chains, while achieving a better likelihood score than SML-PT
50. 

We now select the best performing seeds for SML-PT with 50 chains and SML-APT,
and show in Figure \ref{fig:fup}, the resulting $f_{up}(i)$ curves obtained at
the end of training.

\begin{figure}[ht]
    \centering
    \hspace*{-0.8cm}
    \subfigure[SML-APT]
    {
        \label{fig:swapstat1}
        \includegraphics[scale=0.25]{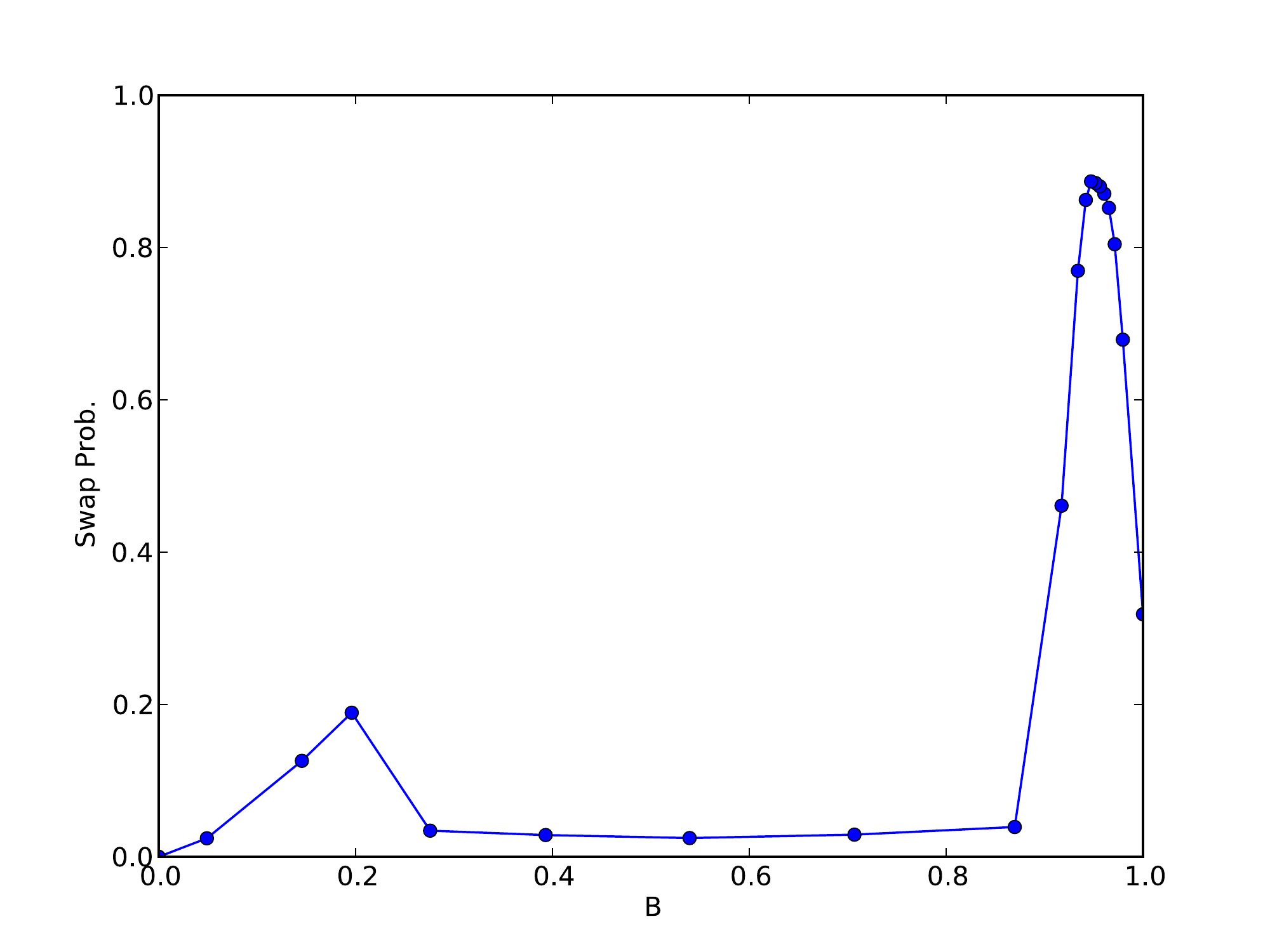}
    }
    \hspace*{-0.8cm}
    \subfigure[SML-PT 20]
    {
        \label{fig:swapstat2}
        \includegraphics[scale=0.25]{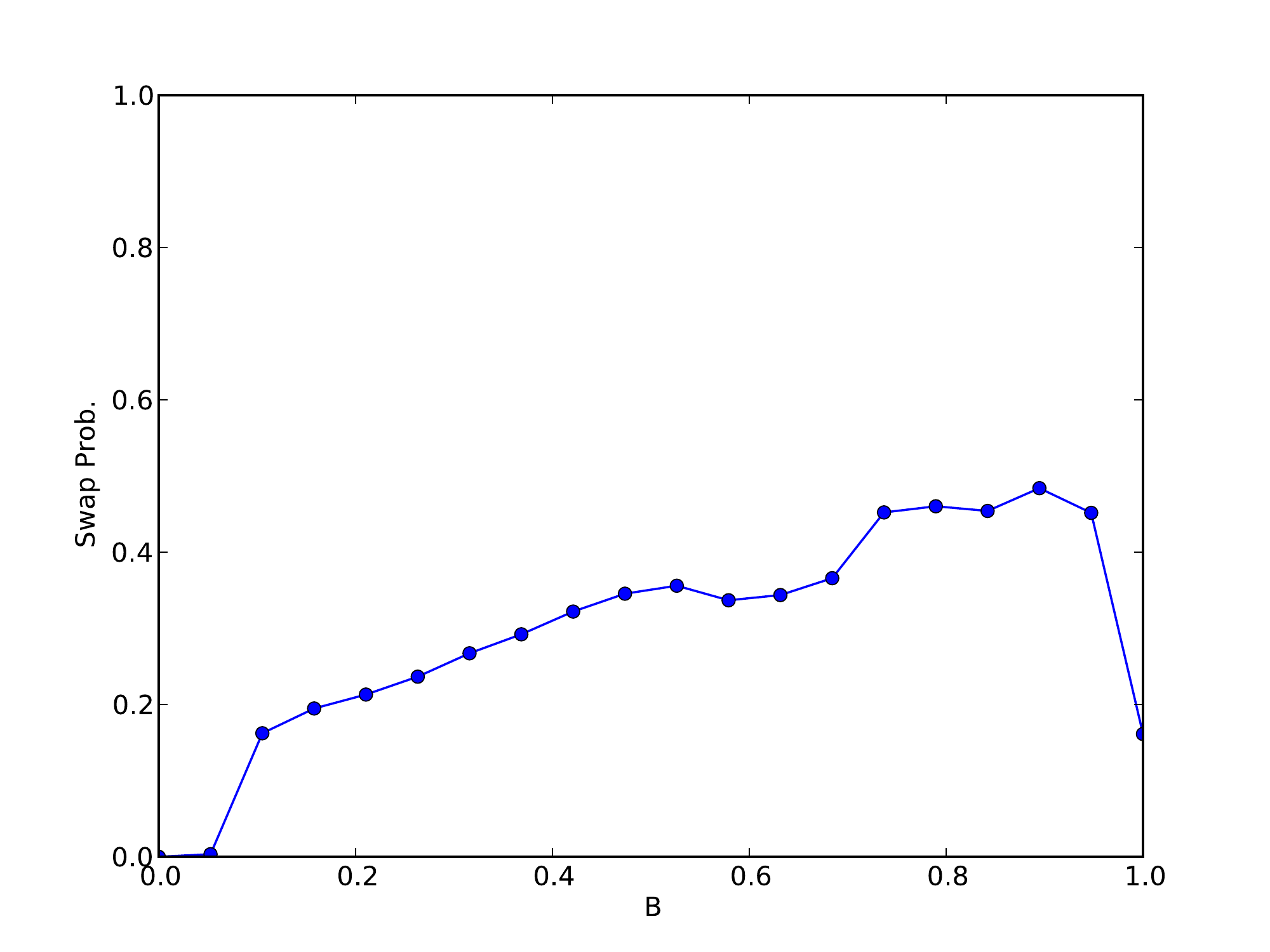}
    }
    \hspace*{-0.8cm}
    \subfigure[SML-PT 50]
    {
        \label{fig:swapstat3}
        \includegraphics[scale=0.25]{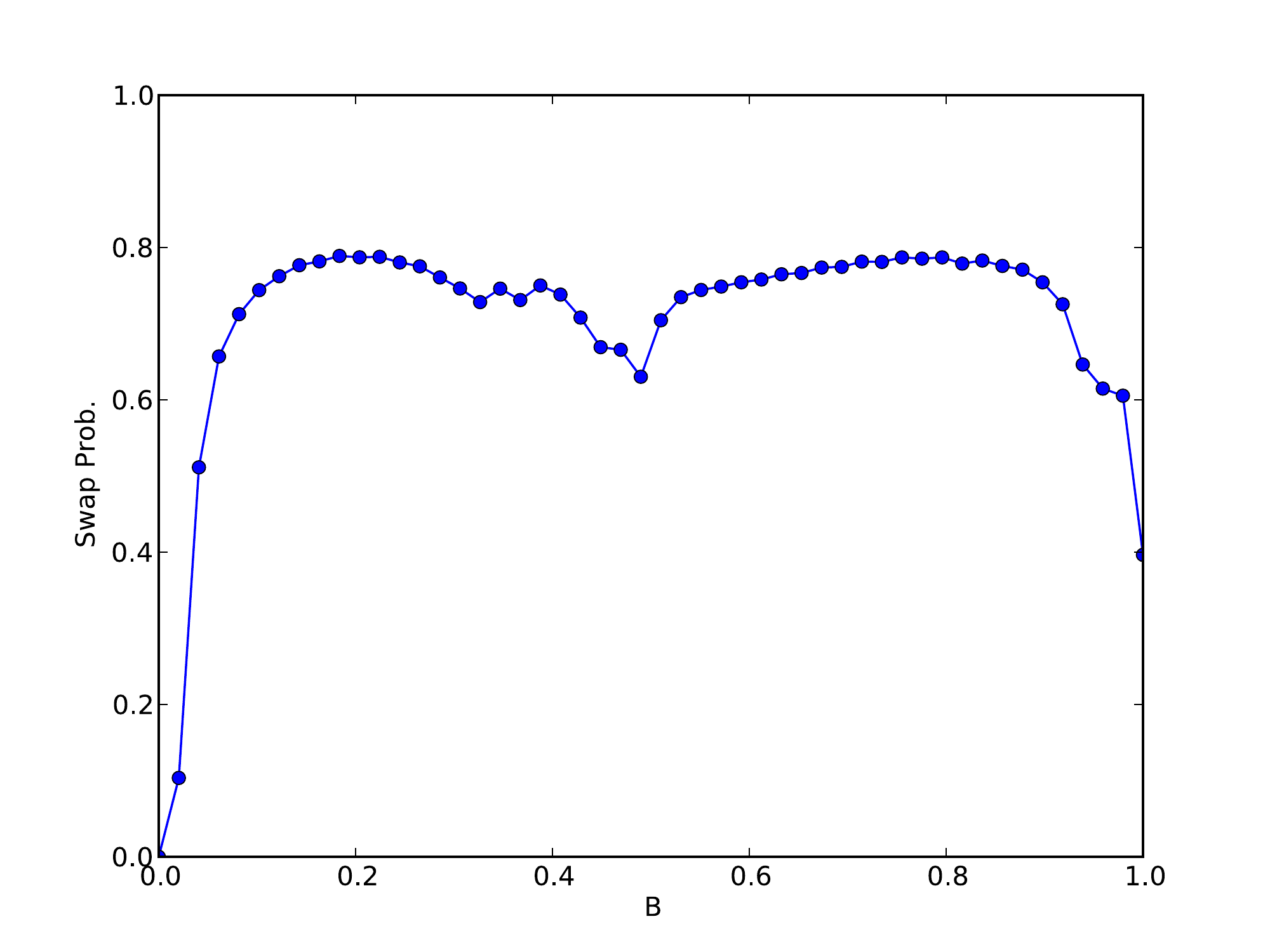}
    }
    \hspace*{-.7cm}
    \caption[] {
    Pairwise swap statistics obtained after $10^5$ updates. Minimizing return
    time causes SML-APT to pool temperatures around bottlenecks, achieving
    large swap rates (0.9) around bottenecks with relatively few chains.
    SML-PT on the other hand results in a much flatter distribution, requiring
    around 50 chains to reach swap rates close to 0.8.
    \label{fig:swapstat}}
\end{figure}

The blue curve plots $f_{up}$ as a function of beta index, while the red curves
plots $f_{up}$ as a function of $\beta$. We can see that SML-APT results in a
more or less linear curve for $f_{up}(i)$, which is not the case for SML-PT. In
Figure \ref{fig:swapstat1} we can see the effect on the pair-wise swap
statistics $r_i$. As reported in \cite{Katzgraber06}, optimizing $\T$
to maintain a linear $f_{up}$ leads to temperatures pooling around the
bottleneck. In comparison, SML-PT fails to capture this phenomenon regardless of whether it uses $20$ or $50$ parallel chains (figures~\ref{fig:swapstat2}-\ref{fig:swapstat3}).

\begin{figure}[ht]
    \centering
    \includegraphics[scale=0.4]{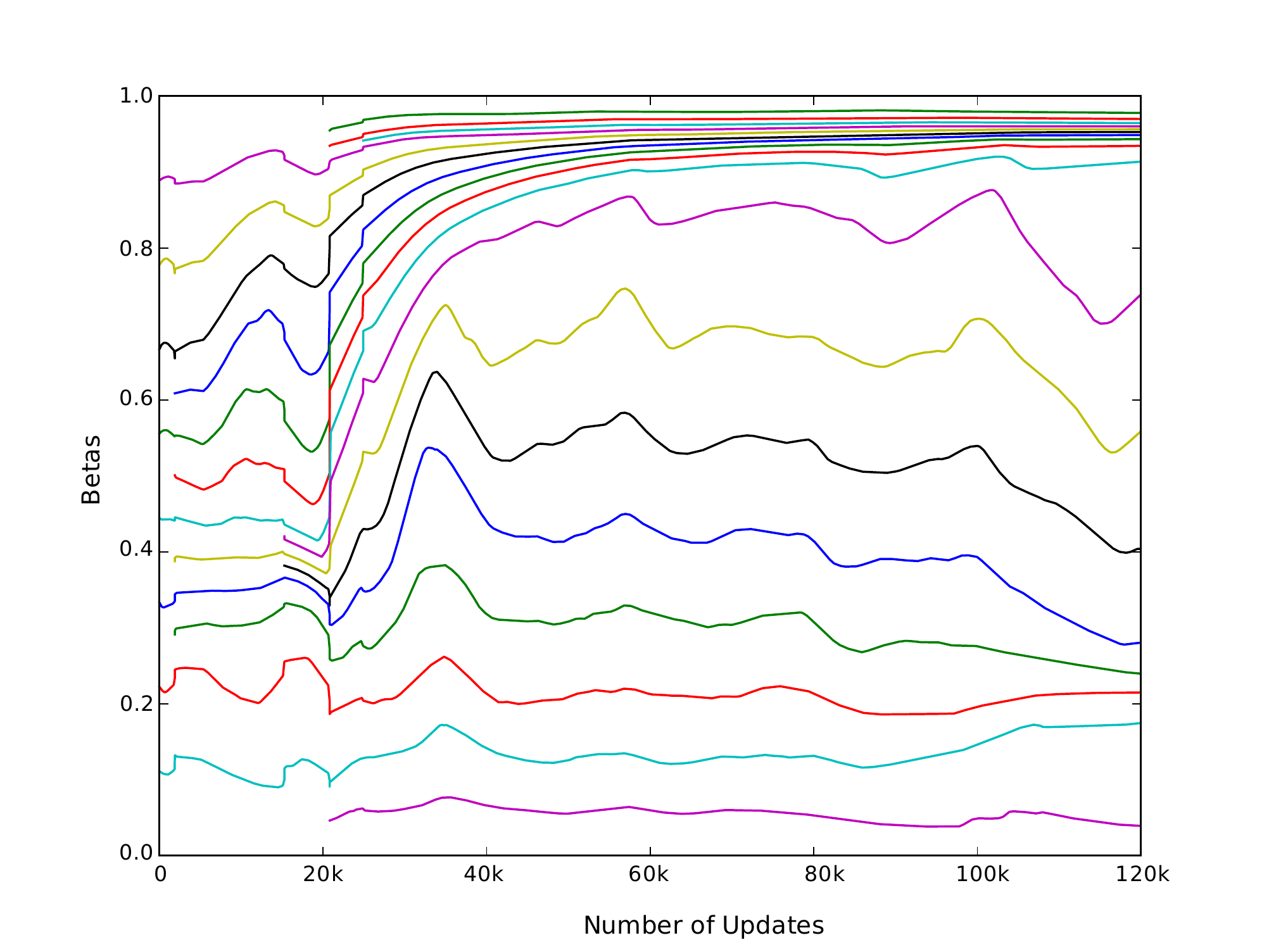}
    \caption{Graphical depiction of the set $\{\beta_i; i \in [1,M]\}$, of
    inverse temperature parameters used by SML-APT during learning.
    Temperatures pool around a bottleneck to minimize return time, while new
    chains are spawned to maintain a given average swap rate.
    Note that the last $20$k updates actually correspond to a pure
    sampling phase (i.e. a learning rate of 0).
    \label{fig:betas}}
\end{figure}

Finally, Figure~\ref{fig:betas} shows the evolution of the inverse temperature
parameters throughout learning. We can see that the position of the bottleneck
in temperature space changes with learning. As such, a manual tuning of
temperatures would be hopeless in achieving optimal return times.

\section{Conclusion}

We have introduced a new adaptive training algorithm for RBMs, which we call
Stochastic Maximum Likelihood with Adaptive Parallel Tempering (SML-APT). It
leverages the benefits of PT in the negative phase of SML, but adapts and
spawns new temperatures so as to minimize return time. The resulting negative
phase sampler thus exhibits greater ergodicity. Using a synthetic dataset, we
have shown that this can directly translate to a better and more stable
likelihood score. In the process, SML-APT also greatly reduces the number of
hyper-parameters to tune: temperature set selection is not only automated, but optimal.
The end-user is left with very few dials: a standard learning rate on $\beta_i$ and a minimum average swap rate $\bar{r}_{min}$ below which to spawn.

Much work still remains. In terms of computational cost,
we would like a model trained with SML-APT and resulting in $M$ chains, to
always be upper-bounded by SML-PT initialized with $M$ chains. Obviously, the
above experiments should also be repeated with larger RBMs on natural datasets,
such as MNIST or Caltech
Silhouettes.\footnote{http://people.cs.ubc.ca/~bmarlin/data/index.shtml}.

\subsubsection*{Acknowledgments}

The authors acknowledge the support of the following agencies for research
funding and computing support: NSERC, Compute Canada and CIFAR. We would also
like to thank the developers of Theano
\footnote{http://deeplearning.net/software/theano/}, for developing such a
powerful tool for scientific computing, especially gradient-based learning.

%\subsubsection*{References}

\nocite{Nadler07}

\small{
\bibliography{strings,strings-shorter,ml,aigaion-shorter}
\bibliographystyle{mlapa}
}

\end{document}